\documentclass[conference]{IEEEtran}
\usepackage{times}

\usepackage[numbers]{natbib}
\usepackage{multicol}
\usepackage[bookmarks=true]{hyperref}

\usepackage{graphicx}
\usepackage{amsmath, amssymb, amsfonts}
\usepackage{booktabs}
\usepackage[font=small,labelfont=bf]{caption}
\usepackage{subcaption}

\newcommand{\be}{\begin{enumerate}}
\newcommand{\ee}{\end{enumerate}}

\newcommand{\bec}{\begin{enumerate}[noitemsep]}
\newcommand{\eec}{\end{enumerate}}

\newcommand{\bi}{\begin{itemize}}
\newcommand{\ei}{\end{itemize}}

\newcommand{\bic}{
    \begin{itemize}[noitemsep]
}
\newcommand{\eic}{\end{itemize}}

\newcommand{\tem}{\textemdash}

\usepackage[dvipsnames]{xcolor}

\newcommand{\green}[1]{{\leavevmode\color{Green}#1}}

\usepackage{tikz}

\newcommand{\Ac}{\mathcal{A}}

\newcommand{\Ec}{\mathcal{E}}

\newcommand{\Gc}{\mathcal{G}}

\newcommand{\Vc}{\mathcal{V}}

\newcommand{\Xc}{\mathcal{X}}

\newcommand{\Rb}{\mathbb{R}}

\newcommand{\av}{\mathbf{a}}

\newcommand{\ev}{\mathbf{e}}
\newcommand{\fv}{\mathbf{f}}
\newcommand{\gv}{\mathbf{g}}

\newcommand{\sv}{\mathbf{s}}

\newcommand{\vv}{\mathbf{v}}

\newcommand{\xv}{\mathbf{x}}

\newcommand{\Fv}{\mathbf{F}}

\newcommand{\Rv}{\mathbf{R}}

\newcommand{\Xv}{\mathbf{X}}

\usepackage{xspace}

\newcommand{\ourmethod}{PIEGraph\xspace}
\newcommand{\piegraphshort}{PIEGraph\xspace}

\newcommand{\piegraphlong}{\textbf{P}hysically \textbf{I}nformed \textbf{E}quivariant \textbf{Graph} Neural Network\xspace}

\newcommand{\forwardSMS}
{\phi_{\mathsf{SMS}}}

\usepackage[normalem]{ulem}

\usepackage{algorithm}
\usepackage{algorithmicx}
\usepackage[noend]{algpseudocode}

\begin{document}

\title{Learning Equivariant Neural-Augmented Object Dynamics from Few Interactions}


\author{%
Sergio Orozco$^{\ast}$,
Tushar Kusnur$^{\dagger}$,
Brandon May$^{\ddagger}$,
George Konidaris$^{\ast}$,
Laura Herlant$^{\dagger}$\\
$^{\ast}$Brown University, Providence, RI\\
$^{\dagger}$Robotics and AI (RAI) Institute, Cambridge, MA\\
$^{\ddagger}$General Motors
}



%

\maketitle
\begin{abstract}
Learning data-efficient object dynamics models for robotic manipulation remains challenging, especially for deformable objects.
A popular approach is to model objects as sets of 3D particles and learn their motion using graph neural networks. In practice, this is not enough to maintain physical feasibility over long horizons and may require large amounts of interaction data to learn.
We introduce PIEGraph, a novel approach to combining analytical physics and data-driven models to capture object dynamics for both rigid and deformable bodies using limited real-world interaction data.
PIEGraph consists of two components: (1) a \textbf{P}hysically \textbf{I}nformed particle-based analytical model (implemented as a spring--mass system) to enforce physically feasible motion,
and (2) an \textbf{E}quivariant \textbf{Graph} Neural Network with a novel action representation that exploits symmetries in particle interactions to guide the analytical model.
We evaluate PIEGraph in simulation and on robot hardware for reorientation and repositioning tasks with ropes, cloth, stuffed animals and rigid objects. We show that our method enables accurate dynamics prediction and reliable downstream robotic manipulation planning, which outperforms state of the art baselines.
\end{abstract}
\IEEEpeerreviewmaketitle

\section{Introduction}
Humans can readily reason about the physical consequences of their actions.
For example, pushing a bottle sideways from the neck causes it to topple over, while pushing a rope along a table causes it to deform over time.
This intuitive understanding enables efficient goal-directed behavior, and enabling robots to reason similarly about object dynamics is a long-standing goal in robotics.
\begin{figure}[!t]
  \centering
  \includegraphics[width=\linewidth]{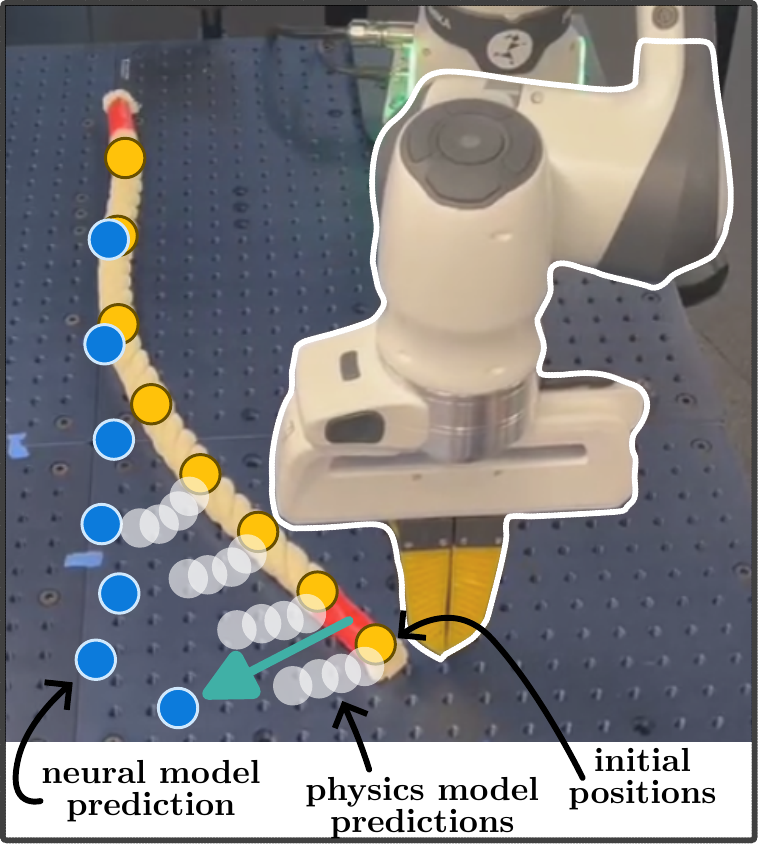}
  \caption{\textbf{General Overview.} We guide physics models toward particle-based neural outputs to guarantee physical plausibility and realistic object motion over long horizons.}
  \label{fig:main}
\end{figure}

Reasoning about the physical consequences of actions is often called an action-conditioned dynamics model.
A model of this form aims to answer the following question: given the current state of the world and a proposed interaction, what is the resultant state?
A dynamics model is useful in robotics for planning, in which future states are used to approximate the cost of robot actions. A dynamics model can also be used in model-based reinforcement learning to improve generalization \cite{young2022benefits, lee2020context} and decrease the amount of training data to train a robot policy \cite{ai2025review,pmlr-v97-hafner19a}.

Dynamics models can be hand-specified for simple cases, such as known rigid objects with uniform density, but for complex scenarios in environments where physical properties are difficult to measure, the dynamics must be learned from data. In this work, we focus on robot manipulation of deformable objects that are pushed, pulled, and lifted from the surface of a table. 

Predicting future states from data is challenging because for many objects and interactions, the physical process is complex and requires a high-dimensional computation. We build on prior work which has identified possible options for representing deformable objects --  including video prediction \cite{b16}, particle-based graph models \cite{b8}, system identification \cite{b1}, and latent-space dynamics learning \cite{b15}. Recent methods \cite{b10, b9} demonstrate that representing objects as sets of 3D particles and modeling their interactions with graph neural networks (GNNs) can outperform many alternative representations, due to the geometric structure and inductive biases provided by particle-based reasoning.

Learning accurate particle dynamics with neural models, however, often requires large amounts of interaction data, frequently involving thousands of robot-object interactions. This is impractical for most robotic applications, so to mitigate this cost, some approaches \cite{b10, b18} rely on simulated data, which introduces the additional challenge of bridging the gap between simulation and real-world behavior. Moreover, without explicit physical structure, learned particle dynamics models may violate physical constraints, for example by drifting, interpenetrating the environment, or failing to preserve object coherence over time.


We introduce \emph{PIEGraph}, a \textbf{P}hysically \textbf{I}nformed particle dynamics model that combines analytical physics with an \textbf{E}quivariant \textbf{Graph} Neural Network (EGNN).
PIEGraph consists of two interacting components: (1) an analytical particle-based physics model, such as a spring--mass system, which enforces short-term physical feasibility, and (2) an action-conditioned EGNN that operates over longer horizons to guide particle motion based on observed data (Figure~\ref{fig:main}).
The EGNN exploits symmetries in particle interactions to improve data efficiency, while the physics model provides an explicit prior that constrains the space of feasible dynamics.
We evaluate PIEGraph on learned dynamics models from human-object interactions, including ropes, cloth, and soft toys, using only a few minutes of interaction data per object.
Our results show that this combination enables accurate dynamics prediction and improves downstream planning performance compared to purely physics-based or learned alternatives.
\section{Background and Problem Setting}\label{sec:background}
\subsection{Robot-Environment Interface and Problem Formulation}
We formulate our problem as a Markov Decision Process (MDP)~\cite{sutton1998reinforcement}.
At time step $t$, the robot constructs a representation of state $x_t \in \Xc$ and, according to policy $\pi$, selects an action $a_t \in \Ac$, where $\Ac$ is a discrete action space.
Conditioned on this action, the dynamics or transition function $\phi$ parameterized by $\theta$ determines the most likely next state
\begin{align}
  x_{t+1} &= \phi(x_t, a_t; \theta) \label{eqn:transition-function}
\end{align}

\textbf{Action parameterization.}
Let an action $a \in \Ac$ have a duration of $\Delta t$ and be parameterized by the 2D coordinates of the robot end-effector at the beginning and end of the action, i.e., $(\sv, \ev) \in \Rb^2 \times \Rb^2$.
We assume that the vector $\overrightarrow{\sv \ev}$ is such that the end-effector stays in contact with the object throughout the duration of the action.

\subsection{Graph Neural Networks for Particle-based Dynamics}\label{subsec:graph-based-dynamics}
Graph Neural Networks (GNNs)~\cite{scarselli2008graph} have proven effective at learning interactions between entities parameterized with neural networks.
Consider a graph $\Gc$ that models state $\Xv$ as 
\begin{align}
    \Xv \triangleq \Gc &= (\Vc, \Ec)  \label{eqn:state-as-graph}
\end{align}
where vertices/nodes $\Vc$ correspond to particles and edges $\Ec$ to pairwise relations among particles \footnote{ignoring time steps for brevity}.
This step typically consists of three basic stages~\cite{sanchez2020learning}:
(1) constructing the graph from the input state, 
(2) a forward pass as multiple rounds of message-passing, and 
(3) extracting the output state from the final graph. 
Let $\phi_{\mathsf{neural}}$ denote a GNN-based dynamics function that performs these stages: $\mathbf{X}_{t+1}^{\mathsf{EGNN}} = \phi_{\mathsf{neural}}(\mathbf{X}_t^{\mathsf{EGNN}}, a_t; \theta)$.

\textbf{Equivariant GNNs.}
The neural dynamics function in \ourmethod is an \textit{Equivariant} GNN~\cite{satorras2021n}.
Besides being invariant to node permutations, EGNNs are additionally translation, rotation and reflection equivariant ($E(n)$) with respect to input nodes. 

The main reason for constraining the neural dynamics function to have equivariant properties is that we are operating in a low-data regime and that we expect the physical motion of the objects to obey the laws of symmetry that equivariance enforces. Namely,  the optimal robot action to take with respect to the object will be equivariant under object rotations and translations \cite{wang2021equivariant}. 

\subsection{Spring-Mass Systems}\label{subsec:spring-mass-systems}
Consider again the graph representation of our particle-based state (Eqn.~\ref{eqn:state-as-graph}).
Let $\forwardSMS$ denote a Spring-Mass System (SMS)-based dynamics function: $\mathbf{X}_{t+1}^{\mathsf{SMS}} = \forwardSMS(\mathbf{X}_t^{\mathsf{SMS}}, a_t; \theta)$.
Let $\xv_i \in \Rb^3$ denote the position, $\vv_i \in \Rb^3$ the velocity, and $m_i \in \Rb$ the mass of the particle $p_i$ corresponding to vertex $v_i \in \Vc$.
An SMS models these particles as connected by springs with stiffness $k_{ij}$  and damping $\delta_i$ coefficients.
Let $\Fv_i$ be the net force acting on $p_i$, i.e., the sum of interaction forces and any external force,
\begin{align}
    \Fv_{i} &= \sum_{j \in N(i)} \Fv^{\mathsf{spring}}_{ij} + \Fv^{\mathsf{ext}}_{i}  
\end{align}
where $N(i)$ is the neighborhood of node $i$.
Spring forces are computed as $\Fv^{\mathsf{spring}}_{ij} =  k_{ij} (||\xv_i - \xv_j|| - r_{ij}) $ where $r_{ij}$ denotes the rest length of the spring, and $\Fv^{\mathsf{ext}}_{i}$ accounts for gravity and contact.
A simulation step $\mathbf{X}_{t+1} = \forwardSMS(\mathbf{X}_t, \Fv_t)$ consists of particle velocities $\vv_i$ and positions $\xv_i$ updated via first-order Euler integration:
\begin{align}
    & \vv_i^{t+1} = \delta (\vv_i^t + \frac{\Fv_i}{m_i} \Delta t);\ \xv_i^{t+1} = \xv_i^t + \vv_t \Delta t  \label{eqn:sms}
\end{align}
\begin{figure}[t]
\centerline{\includegraphics[scale=0.18]{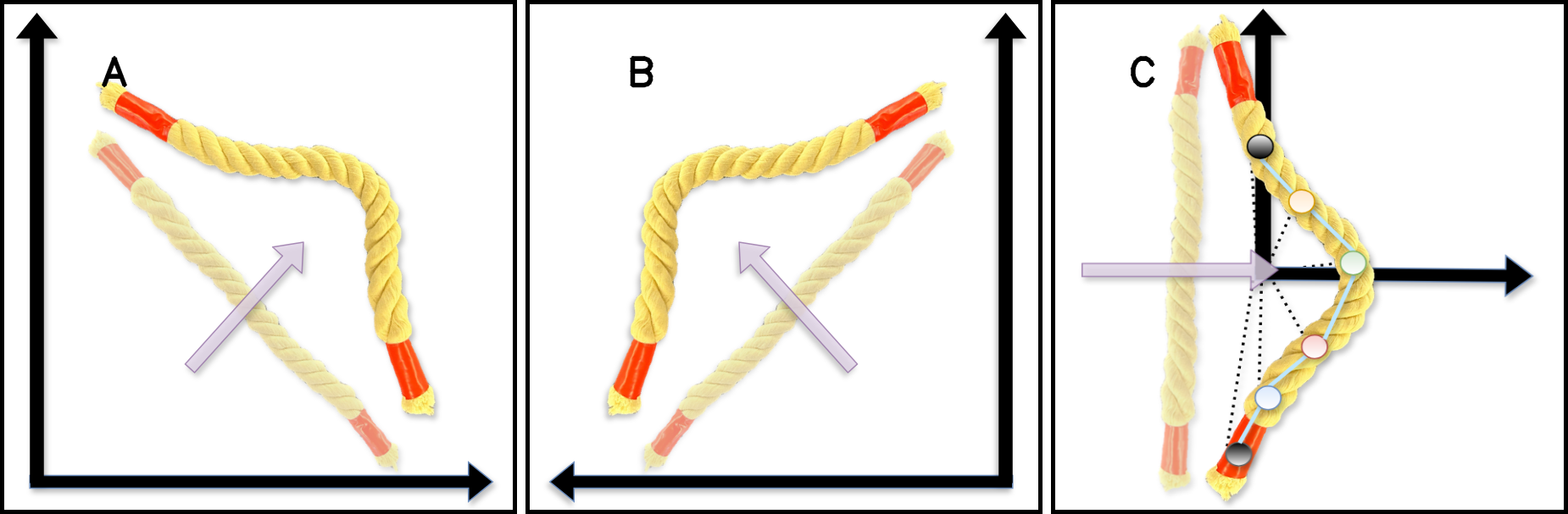}}
\caption{\textbf{Symmetry and Action Canoncialization} \textbf{(a)} We apply a linear push to an object (purple arrow), which results in a u-shaped deformation over time. \textbf{(b)} The same linear push can be applied to the object under some world transformation, meaning that actions in (a) and (b) should be invariant to transformations and canonicalizaed to the object. \textbf{(c)} We align both scenes (a) and (b) to the x axis, such that the actions occur along the x axis. To enforce object pose sensitivity, we calculate the difference between each object particle to the aligned action end position. }
\label{fig:canonical}
\end{figure}
\begin{figure*}[!htbp]
\centerline{\includegraphics[scale=0.8]{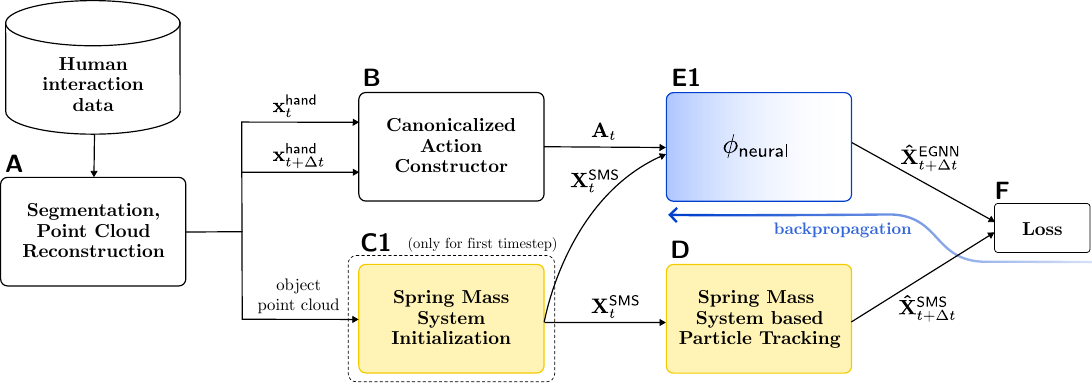}}
    \caption{
        \textbf{Illustrative system diagram\tem training:}
        We train an action-conditioned equivariant graph dynamics model (\textbf{E1}) \green{using MSE loss} (\textbf{F})  from human interaction data captured as an RGBD video.
        We initialize (\textbf{C1}) a spring mass system from an object point cloud at time $t=0$  and track it (\textbf{D}) over multiple actions.
        We track the human's hands at each time step to construct a representation of the action (\textbf{B}).
    }
    \label{fig:training}
\end{figure*}
\section{Related Work}
\subsection{Particle-Based simulation.}
A common dynamics modeling pipeline captures and represents a scene using simulators coupled with system identification on its physical parameters \cite{b6, b1, b5, b3, b4, b2}. While effective, system identification often requires complex, multi-stage optimization over sparse and partially non-differentiable parameters. 
By contrast, our approach does not optimize simulator parameters. Instead, we leverage particle-based simulators to enforce physical feasibility constraints such as object connectivity, shape preservation, collisions, and gravity.
Recent work has also explored constructing object-centric digital twins \cite{b1} using reconstruction techniques such as Gaussian Splatting \cite{b7} and modeling them as particles. 
These methods emphasize accurate 3D reconstruction and tracking, while robotic manipulation is treated as a secondary application with limited quantitative evaluation on their utility for manipulation planning.
We focus explicitly on dynamics learning for robotic planning and find that constructing scenes using initial segmented point clouds and learning the dynamics over that is sufficient for our experiments. 
\subsection{Neural-Based simulation.} A large body of work \cite{b9, b10, b11, b12} models object dynamics through particle-based motion using graph neural networks.
These approaches have demonstrated success on deformable objects including ropes \cite{b10, b13}, cloth \cite{b14}, granular piles \cite{b10}, and stuffed toys \cite{b9}.
Learning accurate dynamics, however, typically requires thousands of interaction trajectories with long data collection times.
This may not be hard to collect in simulation, but bridging the resulting sim-to-real gap often requires additional residual models \cite{b12} or material-specific parameter idenification on material-adaptive models \cite{b10}. 
Our approach differs by explicitly exploiting symmetries in object motion through E(n) Graph Neural Networks \cite{satorras2021n}, enabling data efficient learning directly from limited real-world interaction data.
\subsection{Neural-Augmented Simulation.}
Hybrid approaches that augment physics simulators with neural networks have been explored for tasks such as planar pushing \cite{b21}, object throwing \cite{b22}, and soft robotics \cite{b23}.
While effective, these methods typically target specific interaction modalities or rigid-body dynamics, and often require large datasets.
For example, SAIN \cite{b21} requires thousands of interactions to learn rigid-body dynamics, and TossBot \cite{b22} relies on simple motion primitives and rigid-body assumptions.
Particle-Grid Neural Dynamics (PGND) \cite{b11} is the state-of-the-art particle dynamics model, which outperforms Graph-Based Neural Dynamics\cite{b9}, an extension of Propnet \cite{b8} for real-world deformable object manipulation. PGND is most closely related to our work, as it incorporates physics priors to stabilize particle-based prediction.
Their model predicts velocity fields on an Eulerian grid which are interpolated back to particles via grid-to-point transfers.
By contrast, PIEGraph predicts per-particle displacements and uses PID-controlled forces to guide a spring-mass system toward neural predictions.
Additionally, PGND does not explicitly enforce equivariance, while PIEGraph leverages equivariant architectures to exploit symmetries in object motion.
To our knowledge, no existing neural-augmented simulation method combines particle-based physics, action-conditioned equivariant learning, and data-efficient real-world training at the scale we demonstrate.
\subsection{Equivariant Dynamics Models.}
A variety of equivariant architectures have been proposed for learning dynamics \cite{satorras2021n, b30, b31, b32}.
EGNN, Steerable-EGNN, and Sub-EGNN guarantee equivariance under SE(3) transformations, but they are typically applied to uncontrolled dynamical systems such as molecular dynamics or celestial motion.
While actions can be incorporated, they are not natively supported in these frameworks.
SQPDNet \cite{b31} introduces action-conditioned equivariant dynamics for tabletop pushing, but it models objects as collections of rigid superquadrics and predicts pose transformations, limiting its applicability to deformable objects.
In contrast, PIEGraph models object state at the particle level, enabling expressive dynamics modeling for both rigid and deformable bodies.
\section{Neural-Augmented Particle Dynamics} \label{Method}

\begin{figure*}[!htbp]
    \centerline{\includegraphics[scale=0.8]{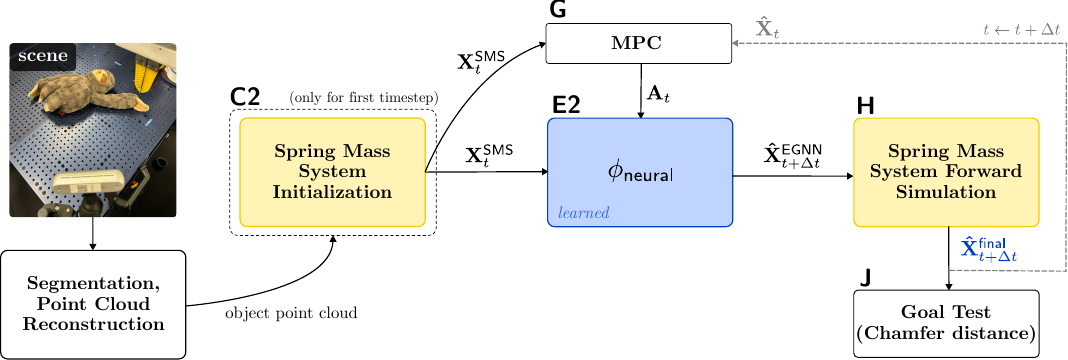}}
    \caption{ 
    \textbf{Illustrative system diagram\tem planning.} We use a learned  equivariant action-conditioned graph dynamics model (\textbf{E2}) to guide (\textbf{H}) a spring mass system constructed (\textbf{C2}) from an initial point cloud of an object at time $t=0$. This guidance process is used to plan (\textbf{G}) for robot actions that reach a specified goal  configuration (\textbf{J}) implemented as a point cloud.}
    \label{fig:planning}
\end{figure*}

\textbf{\piegraphshort}---\piegraphlong---consists of two interacting components:
(1) a particle-based \textbf{physics model} implemented as a spring mass system, and 
(2) an action-conditioned \textbf{neural model} implemented as an E($n$) equivariant graph neural network (EGNN).
The neural model predicts object particle states at the end of an action, providing global targets that guide the physics model.
The physics model then enforces particle-level physical consistency, ensuring that the final predicted state is physically feasible.
Together, these components predict object state transitions with high spatial accuracy while maintaining physical constraints.

\subsection{Physics Model}
$\phi_\mathsf{SMS}$ denotes the spring-mass system update function described in Eqn.~\ref{eqn:sms} and maps the current particle state $\Xv^{\mathsf{SMS}}_t$ and applied forces $\Fv_t$ to the next particle state.
External forces are computed by solving the following tracking objective:
\begin{align}\label{eq:track-force}
\min_{\Fv^{\mathsf{ext}}} \sum_{i}^{N} \sum_{j}^{N}
\left\| \xv^{\mathsf{setpoint}}_{i} -
\phi_{\mathsf{SMS}}(\Xv^{\mathsf{SMS}}_{t}, \Fv^{\mathsf{\mathsf{spring}}}_{j,t},\Fv^{\mathsf{ext}}_t)_i \right\|_2,
\end{align}

where $\Xv^{\mathsf{SMS}}_{i,t}$ denotes particle positions at time $t$, and
$\Fv^{\mathsf{ext}} \in \mathbb{R}^{3 \times n}$ represents per-particle external forces that minimize the distance (Figures~\ref{fig:training}.D and \ref{fig:planning}.H) to a target setpoint configuration $\Xv^{\mathsf{setpoint}}$. We describe  $\Xv^{\mathsf{setpoint}}$ in more detail in Section \ref{method:hierarchy}.
Rather than solving Equation \ref{eq:track-force} in closed form, we approximate the solution using per-particle PID controllers,
\[
\fv^{\mathsf{ext}}_{i,t} = K_{p} e_i(t) + K_{I} \int_0^t e_i(t)\,dt + K_{D} \frac{d e_i(t)}{dt},
\]
where $e_i(t) = \xv_{i}^{\mathsf{setpoint}} - \xv_{i,t}^{\mathsf{SMS}}$, and $K_{p}$, $K_{i}$, and $K_{d}$ are the gain terms which define how the controller reacts to error over time.
These controllers are applied iteratively until the spring--mass system converges to the target configuration.

\subsection{Neural Model}
$\phi_{\mathsf{neural}}$ denotes the neural dynamics update function (Figures~\ref{fig:training}.E1 and \ref{fig:planning}.E2) and maps the current particle state $\Xv^{\mathsf{EGNN}}_t$ and action $\av_t$ to the next particle state.

Instead of using the raw action $\av_t$, defined by the start and end positions of the end-effector, we introduce a canonicalized action representation (Figure \ref{fig:training}.B) that is invariant to global transformations.
Specifically, we decompose $\av_t$ into a start state $\sv_{t}$ and a final state $\ev_{t}$, and define a particle-wise canonical action as
\begin{align}
\av_{i,t}^{\mathsf{canon}} = R^{-(\mathsf{atan2}(\ev_{t} - \sv_{t}) + 2\pi)} (\xv_{i,t} - \ev_{t}),
\quad \forall \xv_{i,t} \in \Xv,
\end{align}
where $\av_{i,t}^{\mathsf{canon}}$ represents the transformation-invariant and canonical action applied to particle $i$. The invariant actions are input into $\phi_{\mathsf{neural}}$ as the node features.
This canonicalization process is illustrated in Figure~\ref{fig:canonical}, and a proof of its transformation invariance is provided in the Supplementary Material. We canonicalize the action-space $a_t$, but not the state-space $X_t$ (shown in equations with a superscript denoting if the particle states came directly from the Spring-Mass System initialization as $X_t^{SMS}$, from the converged Spring-Mass System as $\hat{X}_{t+\Delta t}^{SMS}$, or from the output of the EGNN as $\hat{X}_{t+\Delta t}^{EGNN}$). This is important because there are settings in which state-space canonicalization is difficult or ill-defined, such as in environments with multi-object interactions where no unique canonical frame exists. In these cases, enforcing a canonical representation of the state $X_t$ may require additional assumptions or introduce ambiguities. For this reason, we focus on action-space canonicalization in this work, and use the equivariant structure of the EGNN to take advantage of symmetries in the state-space.

\subsection{Hierarchical Dynamics} \label{method:hierarchy}
While $\phi_{\mathsf{neural}}$ captures long-horizon, action-conditioned object motion, it does not explicitly enforce physical feasibility, and predicted particle configurations may violate constraints such as object shape coherence or ground contact.
Accordingly, during planning we use the neural prediction $\hat \Xv_{t+\Delta t}^{\mathsf{EGNN}}$ as a setpoint that guides the spring--mass system.
Specifically, we set $\Xv^{\mathsf{setpoint}} = \hat \Xv_{t+\Delta t}^{\mathsf{EGNN}}$ in Equation \ref{eq:track-force}, and optimize external forces so that the spring-mass system converges toward the neural prediction while respecting physical constraints such as spring connectivity, rest lengths, and collisions with the environment.
The resulting dual-component prediction is
\begin{align}
    \Xv_{t+\Delta t} =
    \min_{\Fv^{\mathsf{ext}}} \sum_i^N  \sum_j^N
    \left\| \hat \xv^{\mathsf{EGNN}}_{i, t + \Delta t}
    - \phi_{\mathsf{SMS}}(\hat \xv_{t}^\mathsf{SMS}, \Fv^{\mathsf{spring}}_{j,t}, \Fv^{\mathsf{ext}}_t)_i \right\|.
\end{align}

\subsection{Tracking and Data Collection}
For our real-world experiments, we collect video recordings of human-object interactions using four synchronized RGB-D images per frame from different viewpoints at 5\,Hz.
Our dataset consists of 100 interaction sequences (approximately five minutes of data).
We apply a post-processing pipeline using SAM \cite{b19} and MediaPipe Hands \cite{b20} to extract object segmentations and hand trajectories.
At time $t=0$, we construct a spring--mass system $\Xv_{0}^{\mathsf{SMS}}$ from a downsampled point cloud of 50 particles (Figures~\ref{fig:training}.C1 and \ref{fig:planning}.C2), with springs being connected based on hand-specified particle distance thresholds.
Object motion is tracked by iteratively applying external forces, where the setpoint $\xv_i^{\mathsf{setpoint}}$ in Equation \ref{eq:track-force} corresponds to the closest observed point-cloud point to particle $\xv_{i,t}^{\mathsf{SMS}}$ at time $t$ (Figure~\ref{fig:training}.D).
\begin{figure}[t]
    \centerline{\includegraphics[scale=0.35]{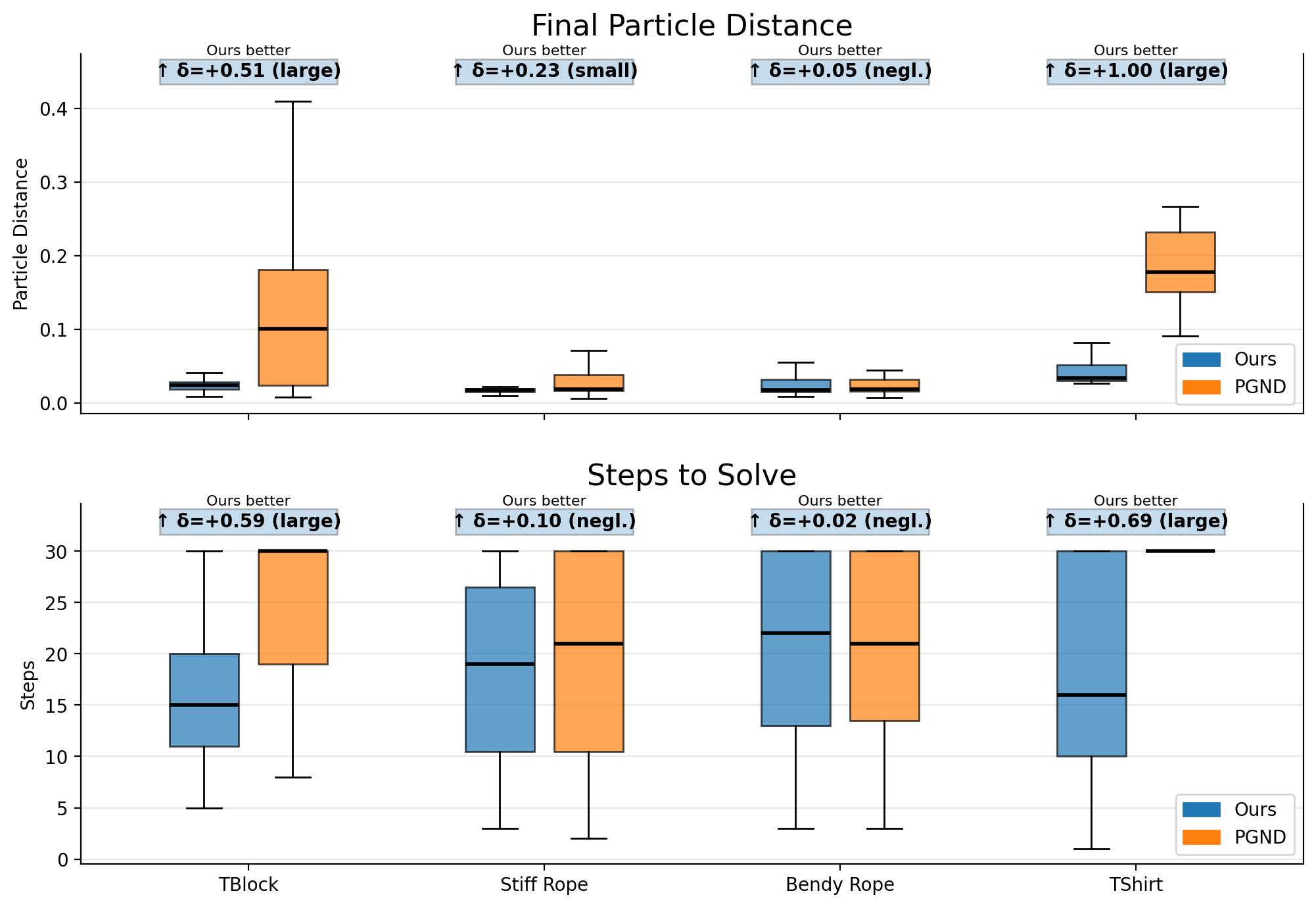}}
    \caption{\textbf{Simulated Planning Results (Quantitative).} For each object, we show the distribution of final state error and steps to solve across all relocation tasks using box plots. Each box displays the 25th percentile, median, and 75th percentile, with whiskers denoting the remaining range while suppressing outliers. Cliff’s delta ($\delta$) is reported for each object to quantify effect size, measuring the probability that a random task solved by our method outperforms the baseline (PGND). Positive $\delta$ indicates improved performance by our method.
    } \label{fig:sim_results_quant}
\end{figure}
\subsection{Training} The neural function $\phi_{\mathsf{neural}}$ is trained using a mean squared error loss between predicted particle positions $\hat\Xv^{\mathsf{EGNN}}_{t+\Delta t}$ and tracked spring-mass states $\hat\Xv^{\mathsf{SMS}}_{t+\Delta t}$, together with a shape consistency loss (\ref{fig:training}.F) that regularizes relative particle displacements:
\begin{align}
L_{\mathsf{dynamics}} = || \hat \Xv_{t+\Delta t}^{\mathsf{EGNN}} - \hat \Xv_{t+\Delta t}^{\mathsf{SMS}} || ^2 _2
\end{align}
\begin{align} \label{eq:loss}
L_{\mathsf{shape}} = \sum_{s \in \mathcal{N}(r) } || (\hat \xv_{t+\Delta t}^{\mathsf{EGNN}, r} - \hat \xv_{t+\Delta t}^{\mathsf{EGNN}, s}) - (\hat \xv_{t+\Delta t}^{\mathsf{SMS}, r} - \hat \xv_{t+\Delta t}^{\mathsf{SMS}, s})||_2^2 
\end{align}
\begin{align}
L = L_{\mathsf{dynamics}} + L_{\mathsf{shape}}
\end{align}
where $\mathcal{N}(r)$ denotes the neighborhood of particle $r$. 

Training data consists of a limited number of prehensile and non-prehensile interactions, which provide only partial coverage of the object’s full interaction space. As a result, evaluation includes object motions and action configurations not explicitly observed during training. Generalization in this setting is facilitated by the use of equivariant representations that exploit symmetries in object motion.
\subsection{Planning}
We perform model-predictive control (MPC) using the Cross Entropy Method (CEM) \cite{b25} to optimize for action sequences.
At each planning step, we sample 500 candidate action trajectories, simulate their outcomes using PIEGraph, and retain the top-performing trajectories over 30 iterations.
Trajectory quality is evaluated based on the distance between predicted particle states and a target goal configuration represented as a fused point cloud (Figure~\ref{fig:planning}.J).
\begin{figure}[t]
    \centerline{\includegraphics[scale=0.5]{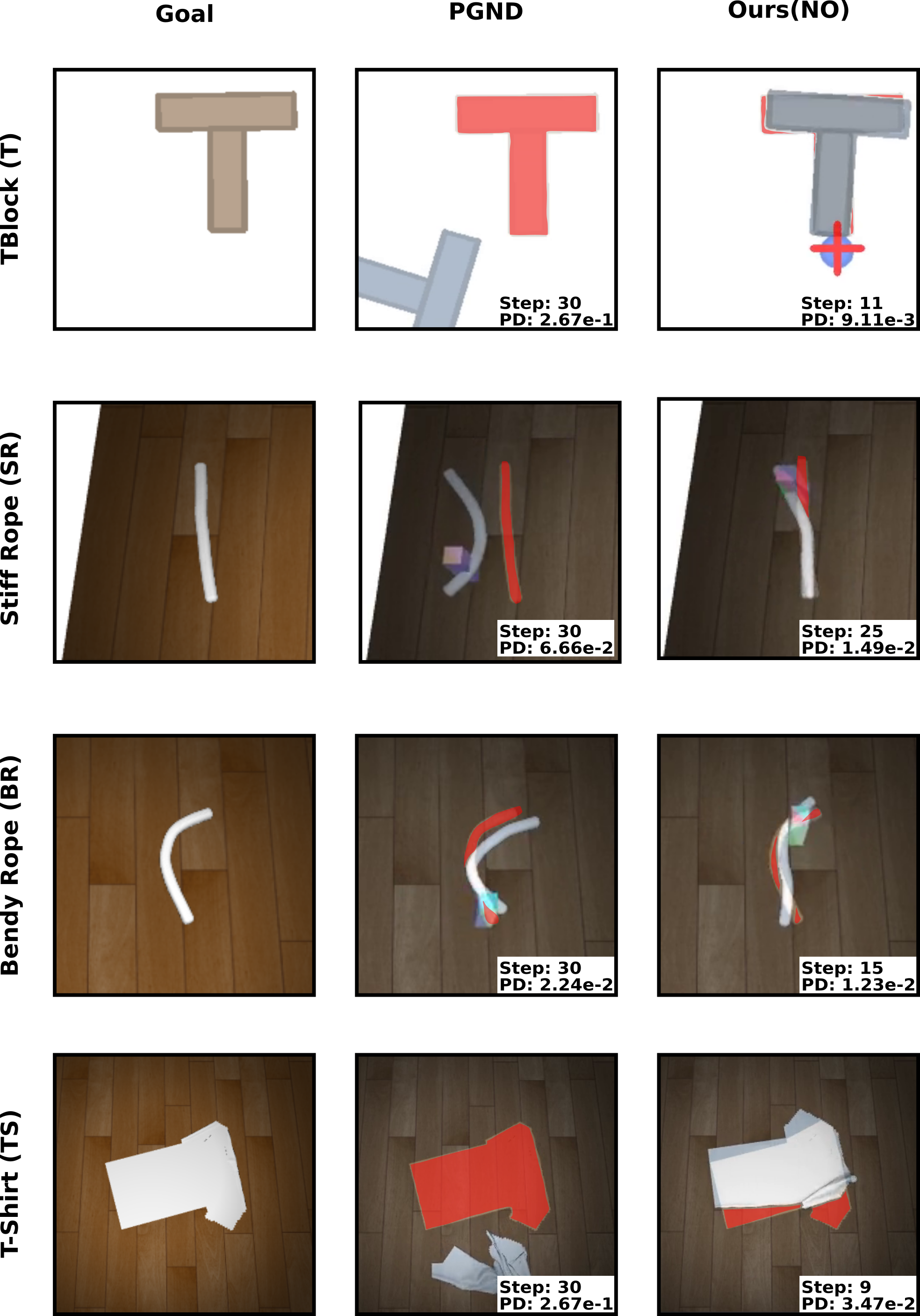}}
    \caption{\textbf{Simulated Planning Results (Qualitative).} For each object, we visualize the goal in the first column and the terminal state after planning using Ours and PGND, respectively. For each terminal state, we manually overlay the goal image in red as a visual indicator for terminal error. We also provide the Particle Distance (PD) value and total steps taken before termination. Values around 0.1 can be safely regarded as failures in planning.}
    \label{fig:sim_results_qual}
\end{figure}

\section{Experiments}

We evaluate PIEGraph in both simulated environments and real-world robotic tabletop settings.
Simulation experiments isolate dynamics prediction and planning behavior under controlled conditions, while real-world experiments evaluate performance under sensing noise, contact variability, and model mismatch.
Our experiments are designed to answer the following questions:

\begin{itemize}
    \item How well does PIEGraph support manipulation planning compared to neural, neural-augmented, and analytical baselines?
    \item What data scales are sufficient for effective planning?
    \item How do architectural choices such as guidance, equivariance, and action representation affect performance?
\end{itemize}

\subsection{Simulated Experiments}
We evaluate in two simulation platforms, namely DaXBench \cite{b33}, an MPM-based simulator for deformable object manipulation, and Pymunk \cite{pymunk}, a 2D rigid-body physics engine.

\textbf{Experiment Setup.}
We perform data collection and evaluation using \emph{pusher} and \emph{picker} agents.
Pusher agents move along the tabletop and interact via collisions, while picker agents grasp objects and translate them along the surface.
Objects include a TBlock, Stiff Rope, Bendy Rope, and a T-Shirt.
For model learning, we collect random agent-object interaction sequences. The number of interactions was chosen based on the complexity of the task as follows:

\begin{itemize}
  \item \textit{TBlock}: A pusher agent manipulates the TBlock, producing unexpected rotations due to its asymmetric center of mass. 10 interaction sequences are used for training.
  \item \textit{Rope}: A pusher agent manipulates stiff and bendy ropes, differentiated by Young’s modulus in DaXBench. 20 and 40 interactions are used for the stiff and bendy rope, respectively.
  \item \textit{TShirt}: A picker agent grasps and moves the cloth, producing large deformations including folds and self-contacts. 20 interactions are used.
\end{itemize}

\textbf{Baselines.}
We compare against Particle Grid Neural Dynamics (PGND), a SoTA particle-based neural dynamics model that incorporates physics priors to stabilize long-horizon prediction. This comparison helps assess the impacts of equivariance and action representation against a similar neural-augmented method in small data regimes.

\begin{figure*}[h]
    \centering
    \includegraphics[width=0.85\textwidth]{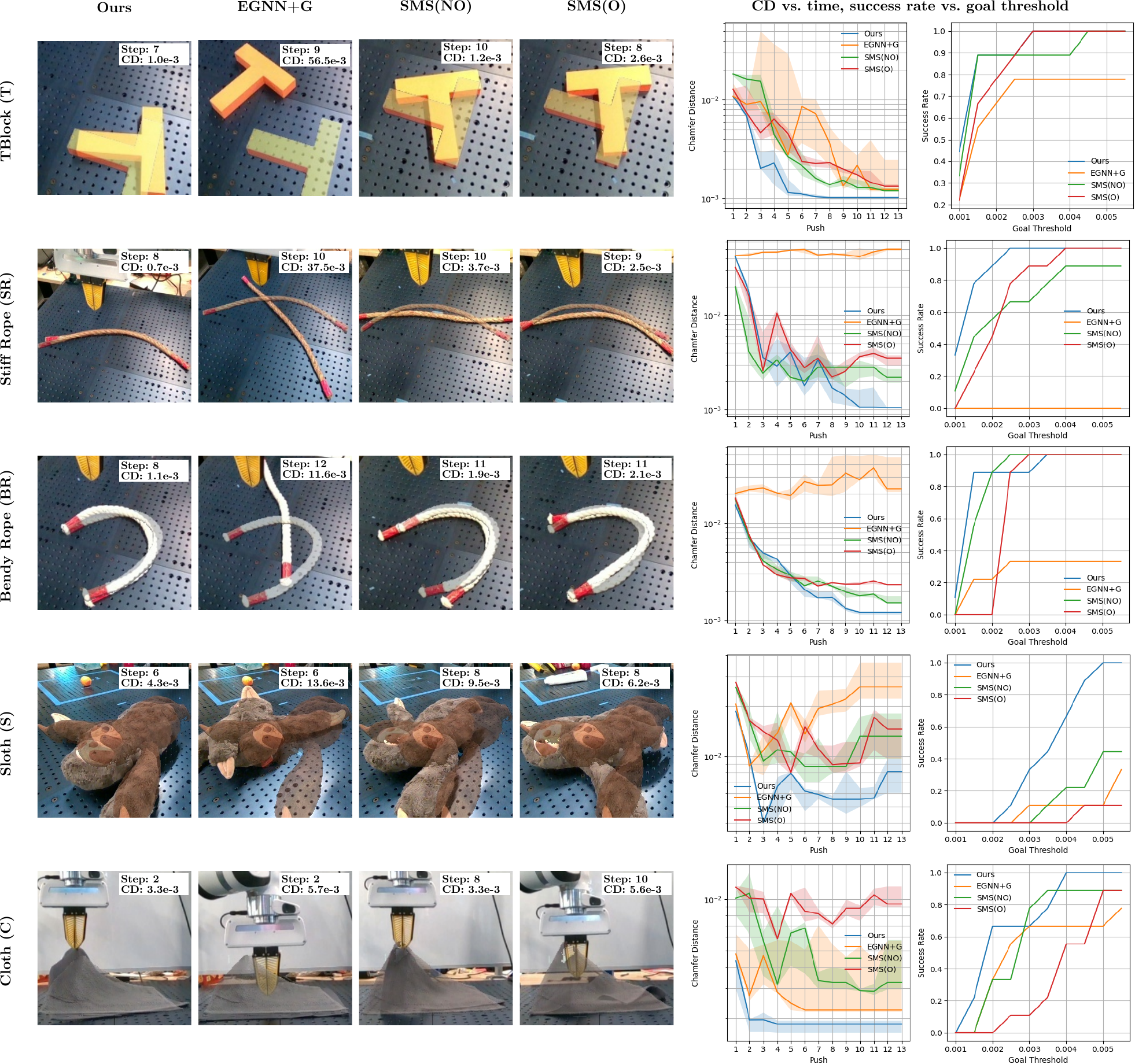}
    \caption{\textbf{Robot Planning Results.} For each object, we plan action sequences to reach a goal configuration implemented as a point cloud. We plan using MPC for 3 separate goals and repeat each experiment 3 times. On the left, we hand-selected qualitative planning results. On the right, we visualize the chamfer distance over time (up to 13 pushes) and task success with varying goal thresholds. We also display the 40th and 60th percentiles as shaded regions to capture variance in performance.
    }
    \label{fig:results}
\end{figure*}

\textbf{Manipulation Planning Results.}
Given a goal configuration $\gv \in \mathbb{R}^{3 \times n}$ represented as a particle set, we plan using either PGND or PIEGraph.
For each object, we define 10 goal configurations and execute up to 30 planned actions.
Each model is evaluated 10 times per goal using different random seeds, which vary the starting location of the object, totaling 100 seeded runs per object.
Planning performance is measured using \emph{Particle Distance}, defined as the mean Euclidean distance between terminal and goal particle configurations.

PIEGraph achieves lower terminal particle distance (and steps to completion) than PGND for the TBlock, Stiff Rope, and TShirt (Figure~\ref{fig:sim_results_quant}). We see negligible performance increase for the bendy rope experiments.
Qualitative inspection reveals that PGND frequently selects suboptimal contact interactions that fail to induce intended object motions.
For rigid objects such as the TBlock, PGND often produces contact forces that induce rotation without sufficient translation, leading to longer or incomplete plans.
For deformable objects, particularly cloth, PGND tends to apply actions that causes excessive local deformation, resulting in bunching and preventing the cloth from flattening toward the goal configuration.
In contrast, PIEGraph selects interactions that more reliably translate the object towards the target configuration, yielding smoother and more goal-directed plans.

To quantify these improvements beyond summary statistics, we report Cliff’s Delta,
$\delta = P(x_{\text{ours}} < x_{\text{pgnd}}) - P(x_{\text{ours}} > x_{\text{pgnd}})$,
where $x_{\text{ours}}$ and $x_{\text{pgnd}}$ denote samples of terminal particle distance (or steps to completion) from PIEGraph and PGND, respectively.
Positive $\delta$ values indicate that PIEGraph outperforms PGND in a majority of seeded trials, confirming that the observed gains reflect consistent improvements in planning accuracy and efficiency rather than isolated successes.
We also visualize goal and terminal states in Figure~\ref{fig:sim_results_qual} to contextualize the quantitative results.
In these environments, terminal particle distances above 0.1 typically correspond to task failure, which PGND exceeds more frequently.

\subsection{Real-World Experiments}
We evaluate on a Franka Research 3 robot with a Finray gripper on a 35$\times$40 inch tabletop.
Objects include a TBlock, Stiff Rope, Bendy Rope, Cloth, and a Stuffed Sloth.
For each object, we collect 100 human-object interaction sequences using 4 Intel RealSense D455 cameras, assuming full object observability.

\textbf{Experiment Setup.}
The robot operates in pusher and picker modes and may autonomously switch between them during planning based on the goal configuration.

\begin{itemize}
  \item \textit{TBlock}: Pushing interactions induce coupled rotation and translation.
  \item \textit{Rope}: Pushing interactions generate complex deformations. Stiff ropes and bendy ropes differ in material stiffness.
  \item \textit{Cloth}: Pushing and picking interactions produce folding and self-contact.
  \item \textit{Sloth}: Pushing interactions induce large, highly compliant deformations of a sloth-shaped soft toy.
\end{itemize}

Visualizations of each of these tasks is shown on the left side of Figure~\ref{fig:results}.

\textbf{Ablations.}
We compare PIEGraph to neural, neural-augmented, and analytical ablations.
\textbf{Ours (NG)} removes physics guidance while retaining equivariant dynamics and canonicalized actions.
\textbf{SMS (NO)} uses a spring-mass system, but with non-optimized parameters.
\textbf{EGNN+G} replaces the canonicalized action representation with particle-based action encodings.
\textbf{SMS(O)} optimizes spring stiffness and damping coefficients using first-order gradient descent in simulation using Warp~\cite{b24}. It is trained using the same training data as our method.
Together, these baselines disentangle the roles of guidance, action representation, and physics priors.

\begin{table}[!h]
\centering
\begingroup
\setlength{\tabcolsep}{2pt}
\begin{tabular}{l || *{5}{c} }
    \toprule
    \textbf{H} & \textbf{Tblock} & \textbf{Stiff Rope} & \textbf{Bendy Rope} & \textbf{Sloth} & \textbf{Cloth} \\
    \midrule
    1 & $(4.1,\; \mathbf{1.3})$ & $(5.1,\; \mathbf{2.5})$ & $(5.5,\; \mathbf{2.4})$ & $(10.9, \mathbf{6.4})$ & $(7.2,\; \mathbf{2.6})$ \\
    2 & $(9.9,\; \mathbf{2.8})$ & $(11.7,\; \mathbf{6.0})$ & $(13.3,\; \mathbf{5.5})$ & $(26.3, \mathbf{14.5})$ & $(15.7,\; \mathbf{5.4})$ \\
    4 & $(23.2,\; \mathbf{5.7})$ & $(27.1,\; \mathbf{12.8})$ & $(31.9,\; \mathbf{13.8})$ & $(63.9, \mathbf{33.7})$ & $(34.6,\; \mathbf{15.7})$ \\
    \bottomrule
\end{tabular}
\caption{\textbf{Dynamics Results.} We present a custom chamfer distance and shape loss metric (\textbf{CD+S}) (See Equation \ref{eq:loss}), for our neural model without and with guidance, respectively --- (Ours(NG), Ours) --- for TBlock, Stiff Rope, Bendy Rope, Sloth, and Cloth for horizon lengths (\textbf{H)} $1, 2, 4$.}
\label{table:1}
\endgroup
\end{table}

\textbf{Dynamics Results.}
Table~\ref{table:1} reports dynamics prediction performance over multiple horizons using the CD+S metric, which combines chamfer distance (dissimilarity between point clouds) and shape consistency loss defined in the second half of Equation \ref{eq:loss}.
While shape regularization improves short-term prediction, Ours (NG) exhibits significant error accumulation over long horizons.
In contrast, guidance through the spring-mass system yields substantially more stable predictions.

\textbf{Robotic Manipulation Results.}
For each object, we define 3 goal configurations and execute up to 13 planned actions.
Each model is evaluated 3 times per goal, for nine total runs per object.
Planning performance is measured using chamfer distance.

Across all objects, PIEGraph achieves higher success rates with fewer planning steps (Figure~\ref{fig:results}).
Comparisons with EGNN+G indicate that the proposed canonicalized action space is critical for learning equivariant dynamics.
Comparisons with SMS(NO) and SMS(O) suggest that purely analytical solutions may struggle to capture subtle object-specific physical effects observed in real-world interactions.  For example, the stiff rope has shape memory that forces it to bend more in one direction. While the simulator baselines fail to model this behavior, PIEGraph properly learns it from interaction data and executes better plans as a result.

\begin{figure}[htbp]
    \centering
    \begin{subfigure}[b]{0.4\textwidth}
        \includegraphics[width=\textwidth]{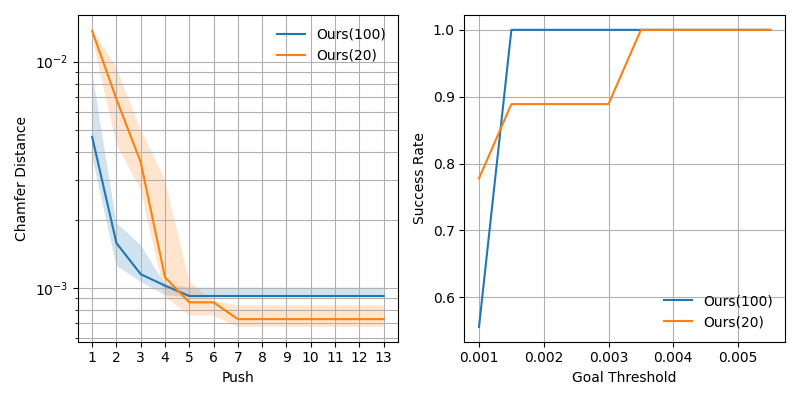}
        \caption{TBlock(20)}
    \end{subfigure}
    \begin{subfigure}[b]{0.4\textwidth}
        \includegraphics[width=\textwidth]{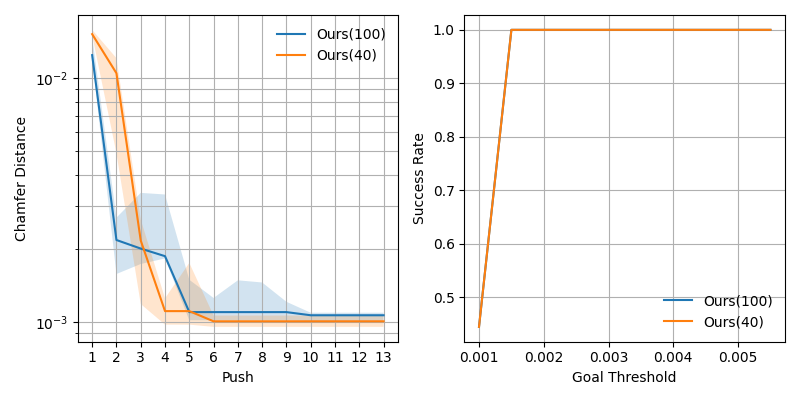}
        \caption{Bendy Rope(40)}
    \end{subfigure}
    \begin{subfigure}[b]{0.4\textwidth}
        \includegraphics[width=\textwidth]{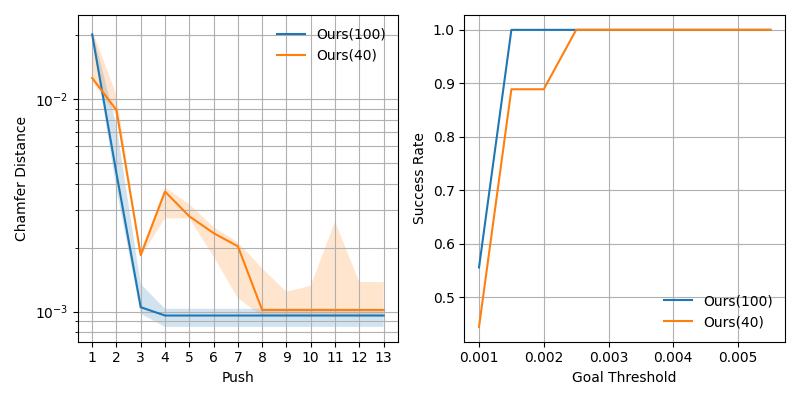}
        \caption{Stiff Rope(40)}
    \end{subfigure}
    \begin{subfigure}[b]{0.4\textwidth}
        \includegraphics[width=\textwidth]{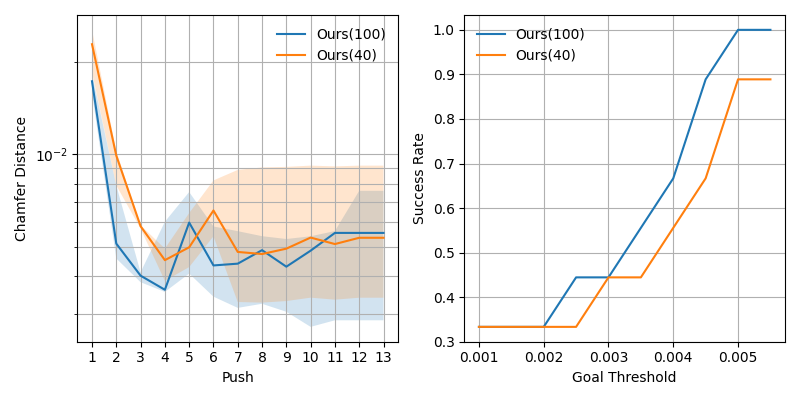}
        \caption{Sloth(40)}
    \end{subfigure}
    \begin{subfigure}[b]{0.4\textwidth}
        \includegraphics[width=\textwidth]{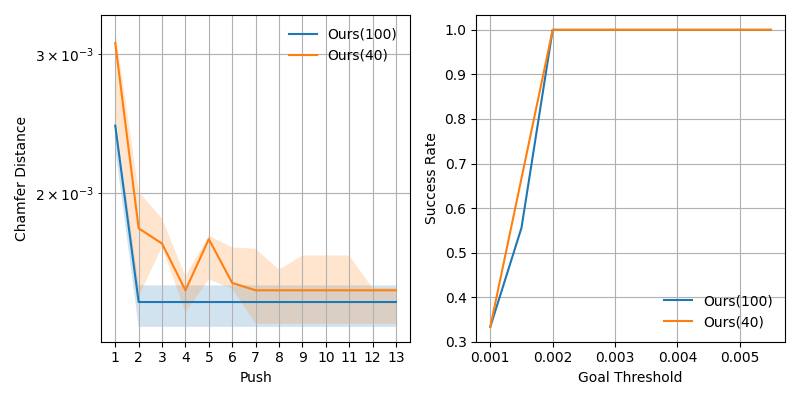}
        \caption{Cloth (40)}
    \end{subfigure}
    \caption{ \textbf{Robot Planning with Different Data Fidelities.}
    We plan actions using MPC for 2 separate goals, and repeat each experiment 3 times. We compare various versions of our models, namely \textbf{Ours(N)} where \textbf{N} is the number of human interactions our model is trained on. For each object we show quantitative results which visualize the chamfer distance over time and task success with varying goal thresholds. We display the 40th and 60th percentiles as shaded regions to capture variance in performance.}
    \label{fig:data-fidelity}
\end{figure}

\textbf{Learning with Different Data Fidelities.}
All real-world experiments use models trained on 100 interaction sequences per object.
We evaluate data efficiency by reducing training data.
Figure~\ref{fig:data-fidelity} shows only modest performance degradation when training data is reduced by more than half.
For the tblock, as few as 20 interactions (approximately one minute of data) remain sufficient for effective planning.
We attribute our model's apparent data efficiency to its exploitation of symmetries in object motion through equivariant representations, as described in Figure \ref{fig:canonical}.

\section{Conclusion}\label{Conclusion}

We presented PIEGraph, a flexible, data-efficient, and physically grounded framework for learning object dynamics from limited real-world interaction data.
By combining equivariant graph neural networks with particle-based physics simulators, PIEGraph enables long-horizon dynamics prediction while maintaining physical feasibility across rigid and deformable objects.
Across both simulated and real-world robotic manipulation tasks, we showed that PIEGraph enables more reliable planning compared to state-of-the-art baselines and mitigates the accumulation of autoregressive prediction errors common in neural dynamics models.
\section{Limitations and Future Work.}
Our experiments assume full observability of object point clouds, and handling partial observations, occlusions, or severe sensing noise remains an important direction for future work.
While PIEGraph can learn dynamics from few human demonstrations, the current formulation primarily targets single-object, non-prehensile manipulation.
This limitation stems in part from the simplicity of the action representation, which models interactions using start and end end-effector positions.
Extending the framework to richer contact modeling and fully three-dimensional prehensile manipulation is a promising avenue for future research.
Additionally, gravity breaks full SE(3) equivariance, which restricts the applicability of standard equivariant architectures in certain manipulation scenarios.
Future work may address this limitation through sub-equivariant graph neural networks \cite{b30}, which explicitly account for gravity.
Finally, our planning evaluation in the real world experiments relies on chamfer distance, which can exhibit geometric ambiguity. Exploring alternative goal representations, such as image-based objectives and particle rendering via Gaussian splatting \cite{b7}, is an exciting direction for extending the framework.

\bibliographystyle{plainnat}
\bibliography{references}

\section{Appendix}

\subsection{Invariant Action Space} \label{Invariant Action Space}
Let $\xv$ be our input state, $\sv$ be our initial end-effector position, and $\ev$ be our final end-effector position. We need to develop an action that is invariant to translations and rotations such that the following statement is true:
\begin{align*}
f(\xv, \sv, \ev) = f(\Rv^\theta \xv + \gv, \Rv^\theta \sv + \gv, \Rv^\theta \ev + \gv) = \av.
\end{align*}
\subsubsection{Proof}
Let's define our action like so:
\begin{align*}
\av &= \Rv^{-(atan2(\ev - \sv) + 2\pi)} (\xv - \ev).
\end{align*}
We need to prove the following equivalence
\begin{align*}
    \av &= \Rv^{-(atan2(\ev - \sv) + 2\pi)} (\xv -\ev) \\
    &= \Rv^{-(atan2(\Rv^\theta \ev + \gv - (\Rv^\theta \sv + \gv) + 2\pi)} (\Rv^\theta \xv + \gv - (\Rv^\theta \ev + \gv)).
\end{align*}
We begin by simplifying,
\begin{align*}
\av = \Rv^{-(atan2(\Rv^\theta (\vv)) + 2\pi)} (\Rv^\theta (\xv - \ev))\textbf{, where } \vv = \ev - \sv.
\end{align*}
We show that $atan2(\Rv^\theta(\vv)) = \theta + atan2(\vv)$ by first converting $\vv$ into polar coordinates like so:
\begin{align*}
\vv = r . \begin{pmatrix}cos(\phi)\\sin(\phi)\end{pmatrix}\textbf{, where } \phi = atan2(\vv).
\end{align*}
Apply $\Rv^\theta$,
\begin{align*}
\Rv^\theta \vv &= r .\Rv^\theta \begin{pmatrix}cos(\phi)\\sin(\phi)\end{pmatrix} \\
&=r \begin{pmatrix}cos(\theta), -sin(\theta)\\sin(\theta), cos(\theta)\end{pmatrix} \begin{pmatrix}cos(\phi)\\sin(\phi)\end{pmatrix} \\
&=r \begin{pmatrix}cos(\theta + \phi)\\sin(\theta + \phi)\end{pmatrix}.
\end{align*}
So,
\begin{align*}
    \Rv^\theta \vv &= r . \begin{pmatrix}cos(\theta + atan2(\vv))\\sin(\theta + atan2(\vv))\end{pmatrix}.
\end{align*}
Thus,
\begin{align*}
    atan2(\Rv^\theta \vv) = \theta + \phi.
\end{align*}
We can now rewrite our action as:
\begin{align*}
\av = \Rv^{-(\theta + \phi + 2\pi)} (\Rv^\theta (\xv - \ev))\textbf{, where } \phi = atan2(\ev - \sv),
\end{align*}
and simplify,
\begin{align*}
\av = \Rv^{-(atan2( \ev - \sv) + 2\pi)} (\xv -\ev).
\end{align*}
$\blacksquare$

\end{document}